\pdfoutput=1

\documentclass[11pt]{article}

\usepackage{acl}

\usepackage{times}
\usepackage{latexsym}

\usepackage[T1]{fontenc}

\usepackage[utf8]{inputenc}

\usepackage{microtype}

\usepackage{csquotes} 
\usepackage{hyphenat} 
\usepackage{expex} 
\usepackage{graphicx} 
\usepackage{booktabs} 
\usepackage{subcaption}
\usepackage{todonotes}
\usepackage{listings}
\title{Word Segmentation and Morphological Parsing for Sanskrit}

\author{Leander Girrbach \and Jingwen Li \\
        University of Tübingen, Germany \\ 
        \texttt{\{leander.girrbach, jingwen.li\}@student.uni-tuebingen.de}
        }

\begin{document}

\maketitle

\begin{abstract}
We describe our participation in the Word Segmentation and Morphological Parsing (WSMP) for Sanskrit hackathon.
We approach the word segmentation task as a sequence labelling task by predicting edit operations from which segmentations are derived.
We approach the morphological analysis task by predicting morphological tags and rules that transform inflected words into their corresponding stems.
Also, we propose an end-to-end trainable pipeline model for joint segmentation and morphological analysis.
Our model performed best in the joint segmentation and analysis subtask (80.018 F1 score) and performed second best in the individual subtasks (segmentation: 96.189 F1 score / analysis: 69.180 F1 score).

Finally, we analyse errors made by our models and suggest future work and possible improvements regarding data and evaluation.

Our code can be accessed from \url{https://github.com/cicl-iscl/TueSan}.

\end{abstract}

\section{Introduction}
\label{sec:intro}
This paper describes our efforts to solve the tasks proposed by the WSMP hackathon\footnote{\url{https://sanskritpanini.github.io/}} associated with Forum for Information Retrieval Evaluation (FIRE) 2021.\footnote{\url{http://fire.irsi.res.in/fire/2021/home}} The hackathon focuses on two Sanskrit NLP problems, namely word segmentation and morphological analysis. These tasks are precursors to downstream NLP tasks like dependency parsing. However, characteristics of the Sanskrit language make them non-trivial and even challenging on their own.

Sanskrit had a rich oral tradition, knowledge and wisdom were passed on through recitation long before the scriptures could be consolidated in written form. It is then not surprising that its orthography faithfully reflects the euphonic modifications made to reduce articulatory efforts during continuous speech or chanting, which requires stricter rhythm. This phonetic smoothing process over consecutive words, termed \textit{sandhi}, comes at the cost of easy and unambiguous identification of word boundaries and word forms. Just by looking at the resulting phonetic stream, which can be written in arbitrary syllabic script or transliteration scheme, it is not enough to directly identify the sequence of words as there exist many lexically and phonetically valid parses. 

Consider the following simple example taken from \citet{guhe2008}, transliterated from Devanāgarī script to IAST:\footnote{International Alphabet of Sanskrit Transliteration}
\vspace{-0.5em}
\pex\label{polygroup}
\begingl
\gla kena pathā bhavānsakhyā sahāgacchat//
\glb kena pathā bhavān{ }sakhyā sah\underline{a{ }ā}gacchat//
\glb kena pathā bhavān{ }sakhyā sah\underline{a{ }a}gacchat//
\endgl
\xe
\vspace{-1.5em}

The italicised first line in~\ref{polygroup} is the sandhied sequence, followed by two segmentations with their difference underlined. Note that the spaces (hiatus) in the original sentence do not fully correspond to word boundaries. 

For someone with knowledge of the language, these are the first two segmentations that come into mind because a human can look at this sentence holistically and first determine the inflected verb hidden in the sandhied chunk \textit{sahāgacchat}. To do so, this person must already have an idea what the sentence is about, combining knowledge of a lexicon, possible sound changes and a grammar. However, a machine can only process the sequence linearly and operate without such linguistic knowledge unless explicitly provided or learned through training. Therefore, determining where sandhi has occurred is already a challenge. But suppose we know for \textit{sahāgacchat}, \textit{ā} is the result of applying the sandhi rule of merging similar word-initial and -final vowels into its long form. Here it is still non-trivial to obtain the word forms because in the reverse process, \textit{ā} can be split into any combination of \textit{ā} and \textit{a}. The possible positions where sandhi could have occurred and the different ways of splitting amount to a multitude of phonetically possible segmentations. Hence the word segmentation task is reformulated as identifying the semantically most valid segmentation for the given sentence \citep{krishna-etal-2017-dataset}. 

For simplicity, here we only discuss the semantically
valid splits for the long vowel \textit{ā}, in particular, \textit{ā} $+$ \textit{a} and \textit{a} $+$ \textit{a}. From the segmentations we obtain the following morphological analyses:\footnote{Glossings used here deviates slightly from the Leipzig Glossing Rules, accessible from: \url{https://www.eva.mpg.de/lingua/resources/glossing-rules.php}. \textsc{par} refers to Parasmaipada}

\vspace{-1em}
\pex\label{poly1}
\a\label{come}
\begingl[everygla=\normalsize\itshape,everyglb=\fontsize{10}{10.8}]
\gla \nogloss{\textit{... sakhyā saha{ }}} \underline{āgacchat}//
\glb a-āgam-t//
\glb \textsc{pst}-\textit{come}-\textsc{3sg.pst.par}// 
\glft '\textbf{\textit{came}} with (a male/female) friend'//
\endgl
\a\label{go}
\begingl[everygla=\normalsize\itshape,everyglb=\fontsize{10}{10.8}]
\gla \nogloss{\textit{... sakhyā saha{ }}} \underline{agacchat}//
\glb a-gam-t//
\glb \textsc{pst}-\textit{go}-\textsc{3sg.pst.par}// 
\glft '\textbf{\textit{went}} with (a male/female) friend'//
\endgl
\xe
\vspace{-2em}

 Without any context,~\ref{come} and~\ref{go} are equally valid. Example~\ref{poly1} also shows that if word segmentation and morphological parsing are performed separately and sequentially, errors from segmentation will propagate into analysis.
 
 Moreover, as \citet{krishna-etal-2020-graph} pointed out, Sanskrit is a language characterised by a high degree of syncretism and homonymy, therefore the morphological analysis for a given surface word form is usually not unique. The inflected noun \textit{sakhyā} in our example sentence, for instance, obtains the following analyses:
\vspace{-1em}
\pex<poly2>
\a\label{masc}
\begingl[everygla=\normalsize\itshape,everyglb=\fontsize{10}{10.8}]
\gla \nogloss{\textit{...}} \underline{sakhyā} saha ...//
\glb sakhi-ā//
\glb \textit{friend}\textsc{.m}-\textsc{ins.sg}//
\glft 'with (a male) friend'//

\endgl
\a\label{fem}
\begingl[everygla=\normalsize\itshape,everyglb=\fontsize{10}{10.8}]
\gla \nogloss{\textit{...}} \underline{sakhyā} saha ...//
\glb sakhī-ā//
\glb \textit{friend}\textsc{.f}-\textsc{ins.sg}//
\glft 'with (a female) friend'//
\endgl
\xe
\vspace{-2em}

Since the same word is analysed in~\ref{masc} and~\ref{fem} with two stems, \textit{sakhi} and \textit{sakhī}, we have a case of homonymy. A more extreme case of homonymy is demonstrated by \textit{śi\d{s}yā\d{h}}, which is the nominative plural of the masculine noun \textit{śi\d{s}ya} (meaning \textit{student}) and also the second person singular optative parasmaipada of the verb \textit{śās} (meaning \textit{to command}). From the feminine noun stem \textit{nau} (meaning \textit{ship, vessel}), \textit{nāva\d{h}} is the resulting form for: (a) ablative singular (b) genitive singular (c) nominative plural (d) accusative plural and (e) vocative plural, hence an example of syncretism.

Notice that the above morphological analyses consider just the individual word without contextual information. It does not realise that \textit{bhavān} is actually a form of the honorific pronoun \textit{bhavat} in the second person, so the verb in~\ref{poly1} is actually intended for the second person, conforming to the inflection paradigm for the third person. This may also pose challenges to downstream tasks if morphological tags were to be used as inputs.

For the simple example sentence in~\ref{polygroup}, ambiguity at both segmentation and analysis level already produces four distinct readings.\footnote{See translations in Example~\ref{poly1}} More ambiguities are expected as the length of the input sentence increases. Using this toy example, we also showed that neither word segmentation nor morphological analysis is to be seen as separated from semantics. Although generation of a sentence is deterministic with a given sequence of stems, their associated morphological properties and a set of sandhi rules, the reverse is non-deterministic at every step. This observation probably encourages postponing decisions at task-level and combining more contextual information to jointly resolve ambiguities.

The rest of this paper is structured as follows: Section~\ref{sec:tasks} gives general descriptions of the three tasks of the hackathon, Section~\ref{sec:data} introduces the dataset provided, Section~\ref{sec:translit} introduces our experiment with transliteration schemes, Section~\ref{sec:method} describes our approaches to each of the three tasks in terms of data preprocessing and model structure, Section~\ref{sec:results} outlines our training and tuning process and the results we obtained. Also, Section~\ref{sec:results} contains an extensive error analysis. Finally, in Section~\ref{sec:discussion} we discuss our findings and conclude this paper. 

\section{Task Descriptions}
\label{sec:tasks}
We briefly introduce the three tasks here, for more detailed descriptions please visit the official hackathon website.\footnote{\url{https://sanskritpanini.github.io/tasks.html}}

\subsection{Word segmentation}

For the first task, a sandhied sentence in the form of a single string is given as input, with unambiguous word boundaries marked by spaces. The desired output is a list of words in the sentence in \textit{unsandhied} form, as shown in Ex.~\ref{task1ex}, taken from the training dataset. Different from other languages where tokenization is also non-trivial, like Japanese, in Sanskrit, another layer of complexity is added to retrieve the correct segmentation: Tokenized outputs also have to undergo a reverse process of the euphonic changes. The task, therefore, is two-fold: finding correct word boundaries and retrieving unsandhied word forms.

\begin{table}[h]
    \renewcommand\tablename{Example}
    \begin{lstlisting}
Input:
    bhavati cãtra
Output:
    ['bhavati', 'ca', 'atra']
    \end{lstlisting}
    \vspace{-1em}
    \caption{Input-output pair for task 1}
    \label{task1ex}
\end{table}
\vspace{-1.5em}

\subsection{Morphological parsing}
In the second task, the segmentation ground-truth is given as input. Our goal is to predict the stem and morphological tag for each word. The morphological tags used in the provided dataset follow the standards of the Sanskrit Heritage Reader \citep{goyal-huet-2016}, where depending on the grammatical category of a word, different numbers of labels are combined to form a single tag string. A finite verb receives a morphological tag with at most six labels, corresponding to 1) mood 2) tense 3) conjugation class 4) voice 5) number and 6) person, while a noun receives a triple consisting of 1) case 2) number and 3) gender as tag. Other indeclinables (avyaya) receive their own POS tags as morphological tags.

\begin{table}[h]
    \renewcommand\tablename{Example}
    \begin{lstlisting}
Input:
    bhavati ca atra
Output:
    [('bhū', 'pr. [1] ac. sg. 3'), 
    ('ca', 'conj.'), ('atra', 'adv.')]
    \end{lstlisting}
    \vspace{-0.5em}
    \caption{Input-output pair for task 2}
    \label{task2ex}
\end{table}
\vspace{-0.5em}

\subsection{Combined segmentation and analysis}
The third task combines task 1 and task 2: Given a sandhied sentence as a single string, we predict the segmented unsandhied words with their full morphological analysis: stem and morphological tag. The output is a list of triples (word, stem, tag) for each sentence.

\begin{table}[h]
    \renewcommand\tablename{Example}
    \begin{lstlisting}
Input:
bhavati cãtra
Output:
[('bhavati', 'bhū', 'pr. [1] ac. sg. 3'),
('ca', 'ca', 'conj.'),
('atra', 'atra', 'adv.')]
    \end{lstlisting}
    \vspace{-1em}
    \caption{Input-output pair for task 3}
    \label{task3ex}
\end{table}
\vspace{-2em}

\section{Dataset}
\label{sec:data}
The WSMP dataset provided by the task organizers is separated into training and development sets. The organizers do not release the test set. The training set consists of 90 000 sentences and the development set contains 10 332 sentences. Ground truths for all three tasks are provided alongside the input sentences. Apart from the main datasets, each sentence is also paired with a graphML\footnote{cf. \citet{krishna-etal-2017-dataset} for structural specifications of the graphML files} file which stores all valid parses of the input sentence generated by Sanskrit Heritage Reader \citep{goyal-huet-2016}. These sentences, as described by the organizers, are taken from the Digital Corpus of Sanskrit\footnote{Accessible from: \url{https://github.com/OliverHellwig/sanskrit/tree/master/dcs/data}} (DCS) \citep{hellwig-dcs}, a corpus for digitalised Sanskrit manuscripts of various domains of knowledge and styles.

\section{Transliteration} 
\label{sec:translit}

There have been opposing views on the effect of using different unit representations in language modeling. \citet{huet-segmentation} argues that transliteration is \textit{completely irrelevant} to Sanskrit text processing and the segmentation problem, while \citet{adiga2021automatic}, in a pioneer large-scale study of Sanskrit ASR (Automatic Speech Recognition), favors \textit{phonetic based graphemic representations} over scripts without one-to-one correspondence to phonemes.

For our own investigation, we experimented with two transliteration schemes for all three tasks to see whether the use of different transliteration schemes has influence over model performance. The first one is the given input in IAST, the other is the internal representation\footnote{Character mapping from IAST is defined in: \url{https://github.com/OliverHellwig/sanskrit/blob/master/papers/2018emnlp/code/data_loader.py}} used by Oliver Hellwig for the DCS, which is similar to SLP1.\footnote{Sanskrit Library Phonetic basic encoding scheme} Different from IAST, which uses two Unicode characters to represent diphthongs and aspirated consonants, the second transliteration scheme uses one Unicode character, establishing one-to-one correspondence between transliterated character and phonemic unit. 

\section{Method}
\label{sec:method}
\subsection{Word segmentation}
\label{sec:method:T1}

We formulate the word segmentation task as a sequence labelling problem: Given a character sequence $s = c_1c_2\ldots c_n$, predict exactly 1 edit operation for each character.
An edit operation locally modifies the input sequence. By applying the predicted edit operations to each character of the source sentence, we arrive at the predicted sequence $s' = c'_1c'_2\ldots c'_n$. Then, we retrieve the predicted segmentation of the input sentence by splitting $s'$ at predicted spaces. In this way, word boundaries and unsandhied word form are determined at the same time.

\paragraph{Data preprocessing}
Our data preprocessing has to achieve two goals: (a) extract the necessary edit operations and (b) align the edit operations with the characters.

First, we remove all datapoints where the source sentence is longer than the target sequences, as these represent corrupted datapoints. This affects $\approx$ 1\% of the dataset. Then, we align each source sequence with the respective target sentence by minimising Levenshtein distance. The aligned sequences decompose into aligned and non-aligned chunks. Aligned chunks are transformed into each other by simple copy operations. Non-overlapping chunks are kept as edit rules, which are represented as tuples of ngrams (\textsc{source}, \textsc{target}). These rules denote what source ngram is transformed into which target ngram. Each source character in \textsc{source} gets the corresponding rule as label. For inference, we contract consecutively predicted equal rules. To ensure \textsc{source} contains at least 1 character, we concatenate \textsc{target}s of rules with empty \textsc{source} to the \textsc{target} of the left rule. Finally, we introduce a special rule for the case where we simply insert a space after a character. \textsc{copy} and \textsc{insert-space} rules abstract away from the concrete character, thereby reducing the total number of rules. The sequences of edit operations generated for the training set are used as ground-truth labels to train our model.

In total, we extract 166 rules from the dataset without transliteration and 187 rules when transliterating Sanskrit inputs. An example of the result of our data preprocessing is in Tab.~\ref{tab:T1:example}. 

\begin{table*}
    \centering
    \resizebox{\textwidth}{!}{
    \begin{tabular}{ccccccccccccccccccccccccccc}
         p & ū & r & v & a & s & y & e & t & i & \textvisiblespace & d & i & n & ā & n & t & a & r & a & k & \d{r} & t & a & s & y & a \\
         C & C & C & C & C & C & C & a\textvisiblespace i & C & C & C & C & C & C & a\textvisiblespace a & C & C & C & C & I & C & C & C & C & C & C & C \\
    \end{tabular}
    }
    \caption{Example of our data preprocessing for task 1: Given the input sentence \enquote{pūrvasyeti dināntarak\d{r}tasya} with corresponding ground truth segmentation \enquote{pūrvasya iti dina antara k\d{r}tasya}, we copy unchanged characters and predict Sandhi rules at word boundaries, if necessary. \enquote{C} mean copy and \enquote{I} mean insert space (after character). For example, we segment \enquote{dināntara} by predicting the Sandhi rule ({ā} $\rightarrow$ {a\textvisiblespace a}) for the first \enquote{ā}.}
    \label{tab:T1:example}
\end{table*}

\paragraph{Model}

\begin{figure}
    \centering
    \includegraphics[width=0.87\linewidth]{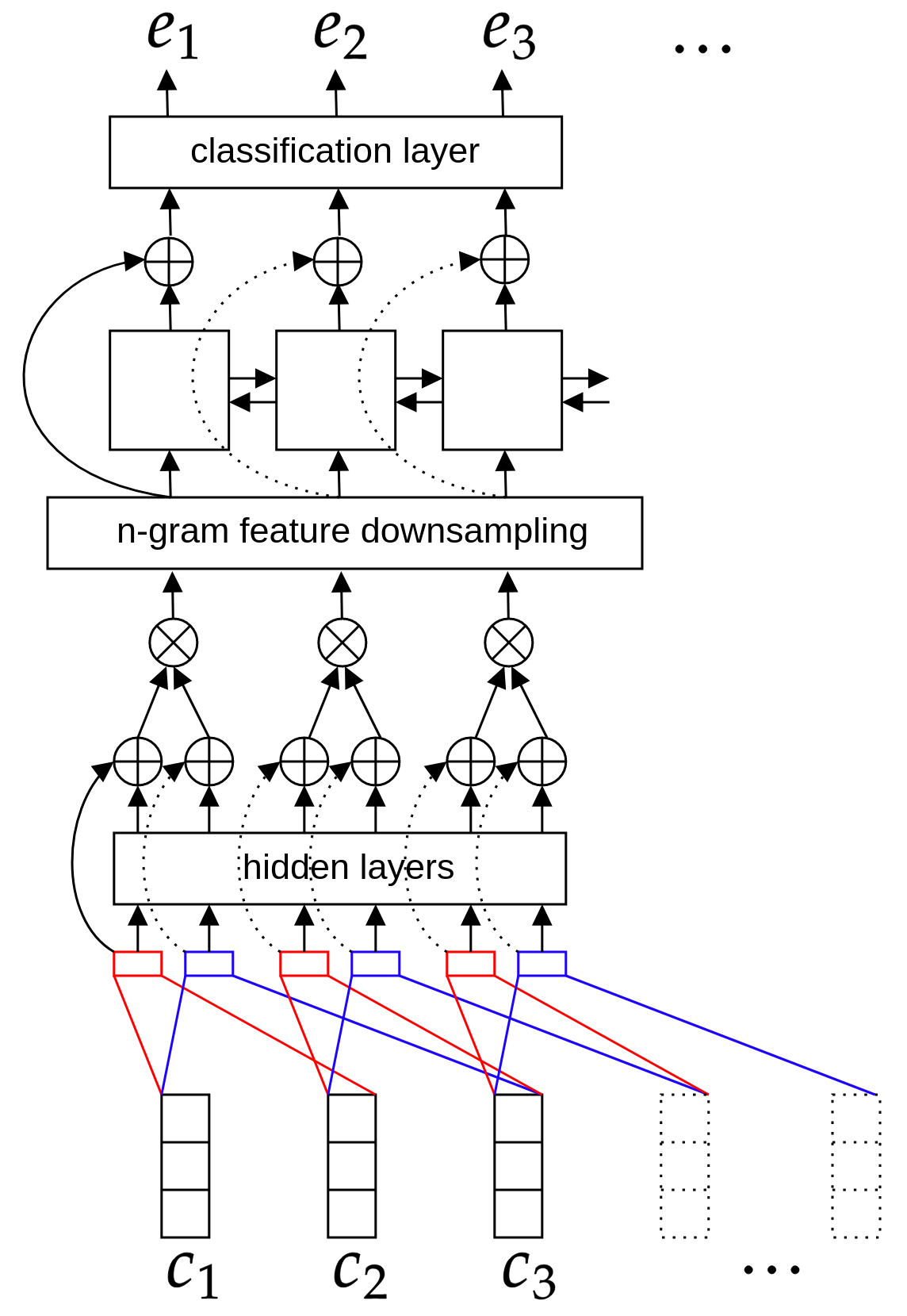}
    \caption{Visualisation of our word segmentation model (Sec.~\ref{sec:method:T1}): Character convolutions of different filter sizes are followed by downsampling and BiLSTM. Finally, rules are predicted for all input characters. Also note the residual connections.}
    \label{fig:T1:model}
\end{figure}

For predicting the label sequence, we use a model that is very similar to the model proposed by \citet{hellwig2018}: First, we extract ngram features from the input character embeddings by 1d-convolutions with filter sizes 2 to 8. Inputs are 0-padded to ensure that sequence lengths remain the same after convolution. Then, we apply a residual block to the ngram features. We concatenate all ngram features and project them to a lower dimension. Finally, we run a 2-layer BiLSTM with residual connection on the sequence and predict the labels for each character. A visualisation of the model is in Fig~\ref{fig:T1:model}.

\subsection{Morphological parsing}
\label{sec:methods:T2}

We also formulate the morphological parsing task as a sequence-labelling task with labels being rules extracted from the train data. These rules describe how to transform a word into the word's stem. Morphological tags are optionally included as part of the rules. If the rules include morphological tags, our model can jointly predict the stem and the tag of a word. Otherwise, the model uses rules only for predicting stems and predicts tags separately. Predicting tags separately turns the task into a multi-task learning problem (cf. \citet{gupta-etal-2020-evaluating}). In general, we find that predicting tags separately yields better performance. One possible reason is that predicting tags separately allows for correct prediction of the tag, even if the wrong rule is predicted for the word. This scenario can arise when the same surface word form has multiple valid morphological analyses.

\paragraph{Dataset preprocessing}

To extract rules, we process the (\textsc{word}, \textsc{stem}) pairs individually. We align \textsc{word} and \textsc{stem} by matching their longest common infix. Then, for both \textsc{word} and \textsc{stem}, we extract a prefix and a suffix. The prefix corresponds to the substring preceding the matched infix. The suffix corresponds to the substring following the matched infix. Both prefix and suffix can be empty. The final rule is thus a quadruple (prefix\textsubscript{\textsc{word}}, suffix\textsubscript{\textsc{word}}, prefix\textsubscript{\textsc{stem}}, suffix\textsubscript{\textsc{stem}}). Given a rule and a word, we can apply the rule if the word starts with prefix\textsubscript{\textsc{word}} and ends with suffix\textsubscript{\textsc{word}}. If this is the case, we can generate the predicted stem by substituting prefix\textsubscript{\textsc{word}} with prefix\textsubscript{\textsc{stem}} and substituting suffix\textsubscript{\textsc{word}} with suffix\textsubscript{\textsc{stem}}.

\begin{figure}
    \centering
    \includegraphics[width=\linewidth]{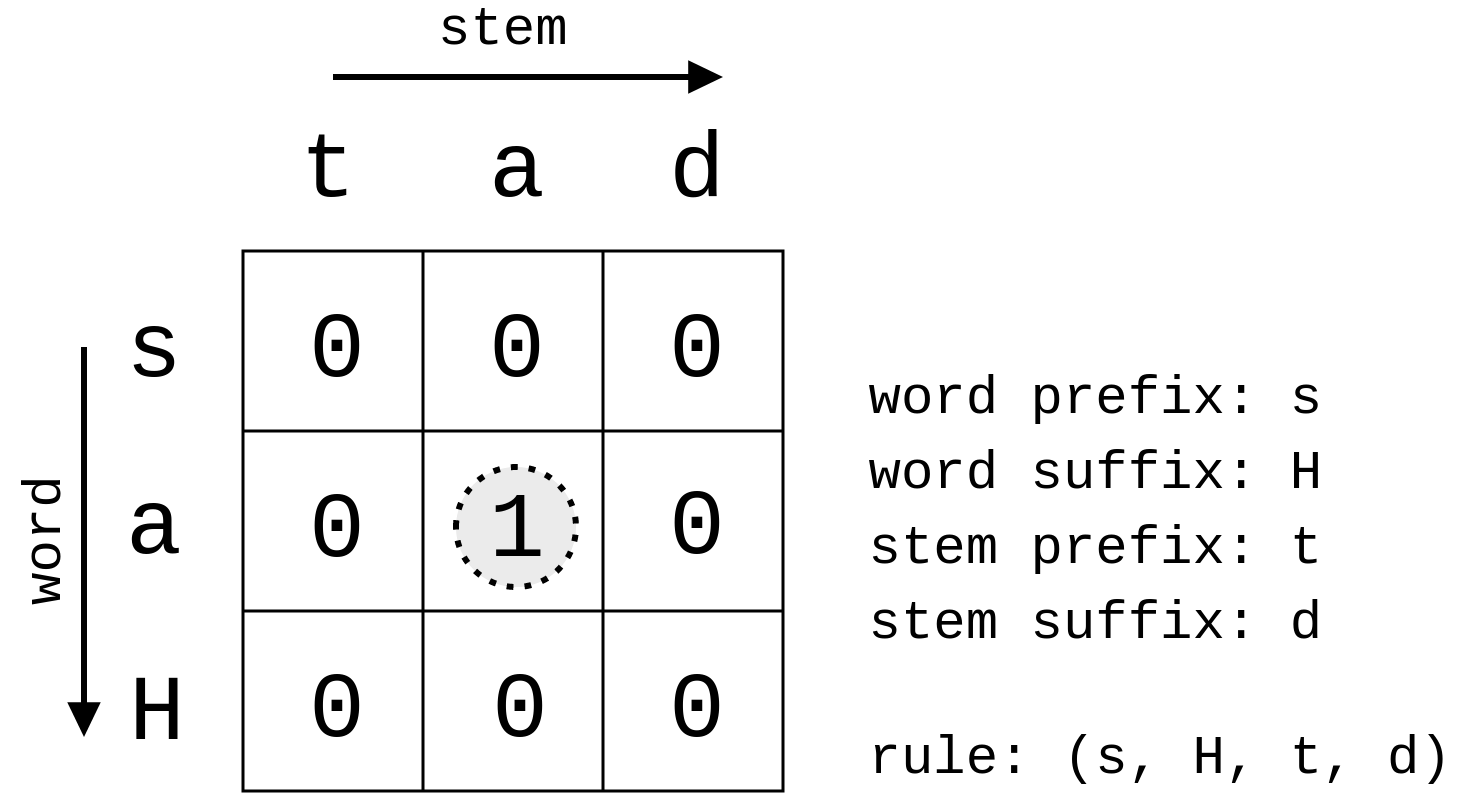}
    \caption{Visualisation of our stem rule extraction method for task 2 (morphological parsing, see Sec.~\ref{sec:methods:T2}): We align the (transliterated) words \enquote{saH} and \enquote{tad}, where \enquote{tad} is the stem of \enquote{saH}, by their longest common infix, \enquote{a}. Then, \enquote{s} becomes prefix\textsubscript{\textsc{word}}
    and \enquote{H} becomes suffix\textsubscript{\textsc{word}}. Likewise, \enquote{t} becomes prefix\textsubscript{\textsc{stem}} and \enquote{d} becomes suffix\textsubscript{\textsc{stem}}.
    }
    \label{fig:T2:alignment}
\end{figure}

An example of alignment and extraction of suffix and prefix is in Fig.~\ref{fig:T2:alignment}.

\paragraph{Model}
For predicting rules (and optionally morphological tags), we first encode each token separately by character convolutions with filter sizes $2$ to $6$. We 0-pad tokens so that their lengths remain unchanged after convolution. The extracted features are combined into a single token embedding by feature-wise max pooling. We also experimented with using the concatenated final hidden states of a \nohyphens{BiLSTM} run on the character features as token embeddings. However, we found this to perform worse.

After having thus calculated an embedding for each token, we run a sentence level BiLSTM on the token embeddings. The reason is that we want to include context information. However, we include a residual connection by adding each token embedding to the respective \nohyphens{BiLSTM} output. In this way, the resulting token embeddings encode both subword and contextual information. Rules and morphological tags are predicted from these contextualised token embeddings.

\begin{figure}
    \centering
    \includegraphics[width=0.8\linewidth]{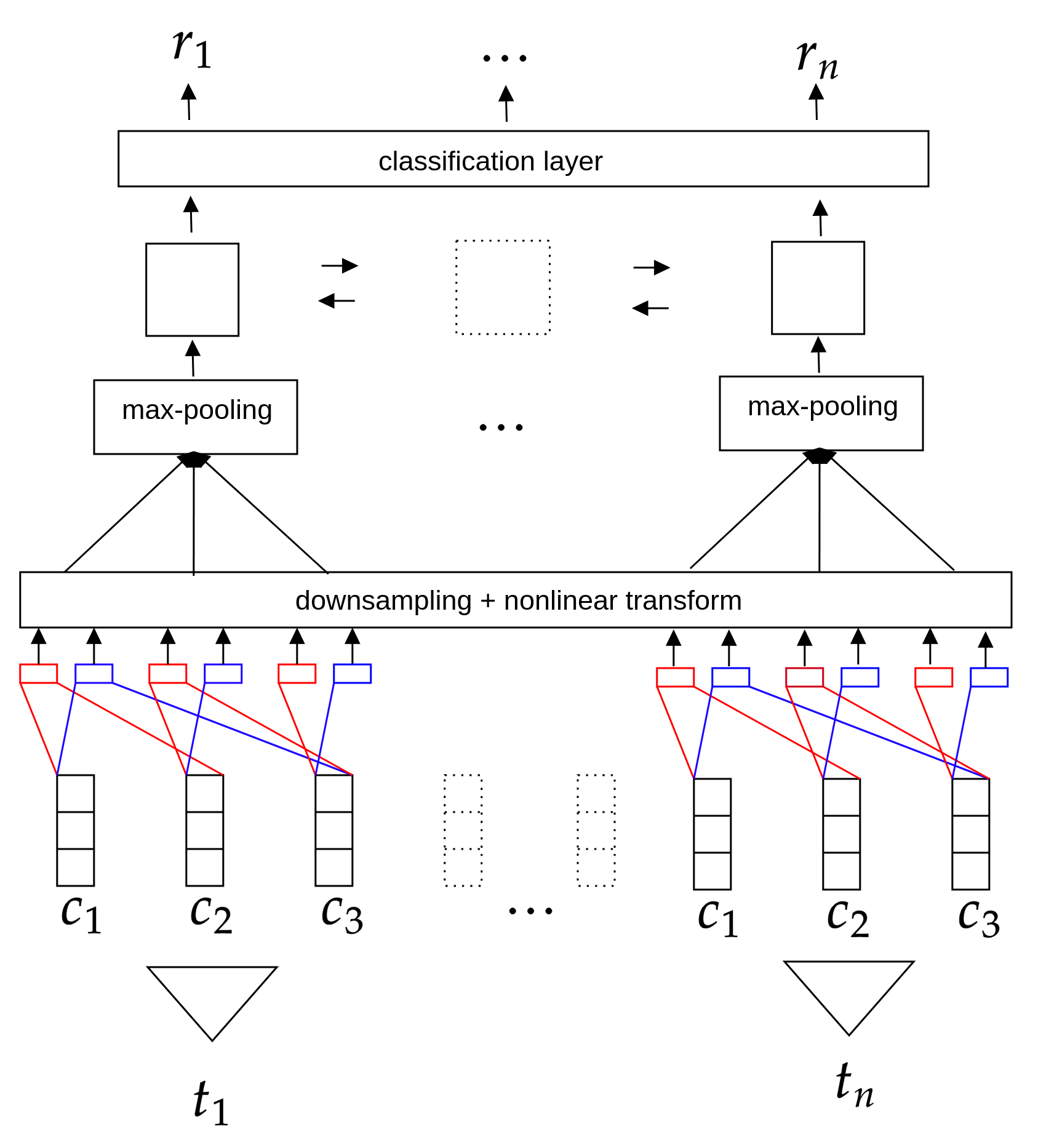}
    \caption{Visualisation of our morphological parsing model (Sec.~\ref{sec:methods:T2}): We encode each word by character convolutions of different filter sizes, which are max-pooled to arrive at embeddings for each word. Then, we capture context information by running a BiLSTM on the token embeddings. Finally, rules are predicted for all input words. Note that residual connections are not shown.}
    \label{fig:T2:model}
\end{figure}

A visualisation of the model is in Fig~\ref{fig:T2:model}. Note that our model can be seen as a modified version of ELMo \citep{peters-etal-2018-deep}.

For inference, we check which rules are applicable to a given token. A rule is applicable to a token iff the token starts with prefix\textsubscript{\textsc{WORD}} and ends with suffix\textsubscript{\textsc{WORD}}. Then, we take the applicable rule with highest prediction probability. If no rules are applicable, we predict the unaltered word.

\subsection{Combined segmentation and analysis}
\label{sec:methods:T3}
We propose a pipeline approach to combined segmentation and analysis that is end-to-end trainable for both predicted segmentations and predicted stems and morphological tags: First, we apply the same model as for word segmentation to predict the words (see Sec.~\ref{sec:method:T1}). Then, we convert the predicted labels into token boundaries ($i, j$), where $i$ denotes the index of the first character of the corresponding word in the unsegmented input string and $j$ denotes the index of the last character. 

For each predicted token, we calculate a token embedding by feature-wise max pooling the corresponding outputs of the BiLSTM model as described in Sec.~\ref{sec:method:T1}. The corresponding outputs are all timesteps of the input sequence starting with index $i$ and ending with index $j$. Finally, we predict stem rules and morphological tags from the token embeddings.

In this case, we always predict stems and morphological tags separately, because this yields better performance in our setup. As noted in Sec.~\ref{sec:methods:T2}, stem rules and morphological tags can be predicted jointly.

Note in particular that we need to predict the segmentation first in order to proceed with the calculation of token embeddings. Therefore, our approach implements a pipeline. However, by reusing hidden states of the BiLSTM, the predictions for stems and morphological tags are differentiable w.r.t. the input characters. Thereofore, the whole model remains end-to-end trainable.

Dataset preprocessing is the same as for task 1 (see Sec.~\ref{sec:method:T1}) and task 2 (see Sec.~\ref{sec:methods:T2}).

\section{Results}
\label{sec:results}
\subsection{Training}
All models are trained by minimising the cross-entropy-loss between prediction probabilities and one-hot encoded target labels. We use a vanilla SGD optimizer without weight decay or momentum. The learning rate is set according to a one cycle learning rate scheduler \cite{smith2019super}. For our models, we found the combination of SGD and one cycle learning rate scheduler to perform better than Adam \cite{DBLP:journals/corr/KingmaB14}. All models are implemented in PyTorch \cite{Paszke2019}.

\subsection{Hyperparameter tuning}
\label{sec:results:hyperparameters}
In order to find the best configuration for our models, we perform extensive hyperparameter tuning using the Ray Tune library \cite{liaw2018tune}.\footnote{\url{https://www.ray.io/ray-tune}} An overview over the tuned hyperparameters and their possible values is in the appendix.

We make the following observations:
\begin{itemize}
    \item Larger embedding sizes always work better. This is surprising, as we even see improvements when the embedding dimension is already larger than the vocabulary size.
    \item Wider models perform better in all evaluated cases. However, this is not surprising given the abundance of training data. We do not evaluate whether deeper models also perform better.
    \item The performance of our models already peaks after 15 epochs. We do not observe improved performance when training longer.
    \item Both the use of transliteration and larger vs. smaller batch sizes do not show a clear trend as to which is better. Instead, this decision is dependent on the particular subtask.
    \item For predicting stem rules for morphological parsing, including all rules generated from the training data (i.e. not discarding infrequent rules) and predicting morphological stems separately from tags yields better performance.
\end{itemize}

\subsection{Hackathon results}
Our approaches obtained the second place at the FIRE 2021 hackathon\footnote{\url{https://sanskritpanini.github.io/index.html}} in both the word segmentation track and the morphological parsing track. Our pipeline model described in Sec.~\ref{sec:methods:T3} obtained the first place in the combined segmentation and analysis track. Since the test set is not released, we report official results from the hackathon's codalab page.\footnote{\url{https://competitions.codalab.org/competitions/35744\#results}, as of \today} Results for all three tracks are in Tab.~\ref{tab:T1:results}, Tab.~\ref{tab:T2:results}, and Tab.~\ref{tab:T3:results}.

\begin{table*}
\begin{subtable}{\linewidth}
\centering
\begin{tabular}{llllll}
\toprule
{} &           & T1 F1 Score & T1 Precision &   T1 Recall & T1 Task Score \\
\midrule
1 &  hrishikeshrt &  97.478 (1) &   97.580 (1) &  97.439 (1) &    97.439 (1) \\
2 &  \textbf{Ours} &  \textbf{96.189 (2)} &   \textbf{96.439 (2)} &  \textbf{96.061 (2)} &    \textbf{96.061 (2)} \\
3 &       jivnesh &  95.436 (3) &   95.793 (3) &  95.237 (3) &    95.237 (3) \\
4 &      maheshak &  82.752 (4) &   82.223 (4) &  83.646 (4) &    83.646 (4) \\
5 &          Asha &  51.470 (5) &   52.177 (5) &  51.525 (5) &    51.525 (5) \\
\bottomrule
\end{tabular}
\caption{Results for T1}
\label{tab:T1:results}
\end{subtable}

\begin{subtable}{\linewidth}
\centering
\begin{tabular}{llllll}
\toprule
{}  &           &  T2 F1-Score &  T2 Precision &   T2 Recall &  T2 Task Score \\ \midrule
1  &       jivnesh &   69.327 (1) &    60.410 (1) &  85.467 (1) &     91.307 (1) \\
2  &     \textbf{Ours} &   \textbf{69.180 (2)} &    \textbf{60.259 (2)} &  \textbf{85.317 (2)} &     \textbf{90.023 (2)} \\
3  &  hrishikeshrt &   42.644 (3) &    52.513 (3) &  39.234 (5) &     48.429 (6) \\
4 &      maheshak &   42.644 (3) &    39.234 (4) &  52.513 (3) &     67.123 (3) \\
5 &          Asha &   38.190 (4) &    33.187 (5) &  47.310 (4) &     62.654 (4) \\
6 &     Manas\_P\_P &   27.950 (5) &    24.017 (6) &  35.384 (6) &     58.265 (5) \\
\bottomrule
\end{tabular}
\caption{Results for T2}
\label{tab:T2:results}
\end{subtable}

\begin{subtable}{\linewidth}
\centering
\begin{tabular}{llllll}
\toprule
{} &           &  T3 F1 Score &  T3 Precision &   T3 Recall &  T3 Task Score \\
\midrule
1 &     \textbf{Ours} &   \textbf{80.018 (1)} &    \textbf{80.252 (1)} &  \textbf{79.906 (1)} &     \textbf{88.404 (1)} \\
2 &       jivnesh &   78.762 (2) &    79.065 (2) &  78.591 (2) &     87.440 (2) \\
3 &  hrishikeshrt &   63.657 (3) &    63.603 (3) &  63.781 (3) &     81.585 (3) \\
4 &      maheshak &   24.026 (4) &    16.579 (4) &  46.403 (4) &     69.606 (4) \\
\bottomrule
\end{tabular}
\caption{Results for T3}
\label{tab:T3:results}
\end{subtable}
\end{table*}

Generally, the top results are very close to each other. Due to the black-box nature of the evaluation, we cannot assess whether all differences are significant. However, we observe that our end-to-end trainable pipeline yields best results for the third track, despite our models do not yield best performance on the individual tasks. One possible conclusion is that end-to-end training in this case makes the pipeline more robust. However, the pipeline can still benefit from our distinct approaches to the two individual tasks.

\subsection{Error analysis}
\label{sec:results:analysis}

Since the test set is not released, we evaluate our model performance on dev data. The scoring script we use is taken from the starting kit provided by the task organisers, which differs from the actual script used for the hackathon\footnote{The updated scoring script can be accessed from \url{https://drive.google.com/drive/folders/1hxeB_t5ymKsiLm9XQuN4gvf_OS4yzRYy}}. 

\paragraph{Segmentation errors}
For word segmentation we report errors of the following types:
\begin{itemize}
    \item Where multiple splits are possible, the model does not pick the correct split. This includes not splitting where it should and oversplitting with the \textsc{insert-space} operation, which is the most common error type. Errors of this type are expected since not all ambiguities can be resolved.
    \item The edit operations extracted are themselves problematic due to incorrect ground-truths. For instance, we extracted rules that map single characters to the empty string, which sometimes causes weird vanishing characters in the prediction. Upon inspection, we found that some of the \textit{ground-truths} are marked with this problem, so we may have extracted such rules from corrupt datapoints in the training data.
    
    
    
    \item We observe false errors due to incorrect ground-truths. The gold segmentations can sometimes be incomplete, either characters or entire chunks of the input sentence are missing. 
    
    
\end{itemize}

\paragraph{Analysis errors}

For predicting stems and morphological tagging, our T2 model evaluated on dev set produces 65.51\% full match for the sentences. Out of the 68 094 words to be predicted, 92.51\% are fully correctly predicted. And 87.69\% of the erroneous cases are partially correct, which means either the stem or the tag is correctly predicted. Note that the tag is a single string combining multiple morphological properties and the evaluation we choose to report here uses strict comparison. The hackathon, however, uses a different way to compare the strings, which will be discussed later in~\ref{sec:results:eval}. 

For morphological tagging, our predictions are wrong most of the time only by one or two subtags. For instance the model predicts \enquote{m. sg. acc.} instead of \enquote{n. sg. acc.} for the form \textit{deham} (meaning \textit{body}), only the gender is incorrect. Since gender is an inherent property of the noun, and according to the declension paradigm of masculine and neuter nouns ending in short vowel \textit{-a}, both analyses are valid without contextual information or a lexicon. Similarly, for verbs, the conjugate class is associated with the root, we observe cases where only the class is incorrectly predicted. For example the verbs in one particular conjugation class, the tenth, are often confused with causatives. They are sometimes indistinguishable for human as well, hence our model predicted \enquote{ca. opt. ac. sg. 3} where the correct analysis would be \enquote{opt. [10] ac. sg. 3}.

For recovering word stems, the model does not perform well when sound changes are involved, in particular the retroflexion of n and s to \d{n} and \d{s}, or one induced by an insertion of a past tense marker \textit{a} between a preverb and the stem, like for \textit{abhidhāv-}, when \textit{a} is inserted between the preverb \textit{abhi-} and the stem \textit{dhāv}, it changes to \textit{ahbyadhāv-}, our model prediction fails to remove the tense marker \textit{a} and restore \textit{i} from \textit{y}.

\subsection{Reevaluating evaluation}
\label{sec:results:eval}
Since the hackathon is essentially a competition, it is necessary to look at how the scores are calculated, a more fine-grained way of evaluation may change the final results depending on the models used by the teams. As mentioned before, the hackathon uses a different way to compare strings. Instead of checking for string equality, they used a counter-based method to count the characters that occurred in both the ground-truth and the model prediction. Although this measure seems more lenient, it is not very accurate and can create more confusion. 

To illustrate the point, consider the made-up example of \enquote{opt. [1] ac. sg. 3} and \enquote{opt. [3] ac. sg. 1}, according to their scoring script, this prediction is considered fully correct, when the third person is predicted as first person\footnote{the conjugate class is also wrong, but is of lesser importance}. On the contrary, \enquote{ca. opt. ac. sg. 3} and 
\enquote{opt. [10] ac. sg. 3} are considered different when they are actually more similar to each other than the previous pair. For nouns, consider \enquote{f. du. abl.} and \enquote{m. pl. dat.}, none of the subtags are correctly predicted, but 8 out of 11 characters are counted as the same\footnote{including dots and spaces}, merely because there happen to be overlapping characters in different morphological categories. This would imply that locative (\textit{loc.}) and vocative (\textit{voc.}) are considered more similar than nominative (\textit{nom.})and vocative (\textit{voc.}), while the truth is the exact opposite. In this regard, this way of evaluation is not very informative.

When models achieve similar high scores, it is also interesting to know on what they fail and whether model performance is consistent over the "difficult" cases. To better understand individual model performance, we propose to subdivide the test dataset into multiple sets, each sharing some linguistically-oriented properties. For instance, an expert in Sanskrit can compose lists of hard-to-analyse forms categorised into forms with reduplication, ambiguous forms, idiosyncratic forms, forms with nontransparent vowel gradation etc. Subsets of the test set can be automatically created from there on. It would also be useful to have a set testing the models on out-of-vocabulary (OOV) forms of known roots and unseen roots. In doing so, we have a more fine-grained qualitative evaluation of the models. The results would be more informative especially when model structures are not specified in the competition.

\section{Discussion and Conclusion}
\label{sec:discussion}

In this paper, we have presented our approaches to Sanskrit word segmentation and morphological analysis.
Our main contributions are proposing to predict rules to generate stems from inflected forms and proposing an end-to-end trainable but pipelined model for joint segmentation and morphological parsing. Since our rules are applicable to any word with matching prefix and suffix, they can be applied to analyse OOV forms, transferring the inflection paradigm we learned to potentially unseen roots. An analogous idea, "data hallucination" \cite{anastasopoulos-neubig-2019-pushing}, is already used for data augmentation in learning morphological inflection in a low-resource setting, where the stem region of a word is identified via alignment and characters in that region are replaced randomly. While they are hallucinating data, we are essentially doing root/stem interpolation. 

Our approaches performed well in the WSMP hackathon associated with FIRE 2021.
This shows that these ideas constitute interesting avenues which can be further explored in the future.

Predicting rules for segmentation and stemming offers the possibility of including linguistic information both into preprocessing (e.g. which rules to consider, the format of the rules) and decoding (often, there are only few rules that apply in to certain instance, which should be narrowed down as much as possible). This is confirmed by our error analysis, where we spot erroneous rules derived from corrupted data and wrong predictions that could be corrected by a dictionary lookup (e.g. wrongly predicted gender).

One line of future work therefore could enhance our approach by including more explicit linguistic information, for example by consulting a dictionary when decoding to rule out stemming rules or tags. Also, we do not make use of the segmentation lattices representing all valid segmentations. These could be used to improve alignments for extracting segmentation edit operations. 

Our error analysis also shows that the models' performance still can be improved. Therefore, investigating the impact of pretrained embeddings and better suited model architectures would be useful.

Finally, we discussed problems with the data quality and evaluation that we think could help improve NLP processing of Sanskrit in the future.

\section*{Acknowledgements}
We thank {\c{C}}a{\u{g}}r{\i} {\c{C}}{\"o}ltekin for support and helpful discussions. We also thank the NVIDIA corporation for donating GPUs to {\c{C}}a{\u{g}}r{\i}, which we used to run our experiments. We thank Frank Köhler for his work in teaching Sanskrit. Finally, we extend our thank to the organisers of the FIRE 2021 hackathon for organising the event and providing data.

\bibliography{custom}
\bibliographystyle{acl_natbib}

\appendix
\section{Appendix}
\label{sec:appendix}
\subsection{Hyperparameters}
As stated in Sec.~\ref{sec:results:hyperparameters}, we perform extensive hyperparameter tuning to find good configurations for our models. The tuned hyperparameters and corresponding values are in Tab.~\ref{tab:hyperparams}.

Due to time and resource restrictions, we only evaluate hyperparameters that seem either interesting from a theoretical perspective (e.g. whether to use transliterated inputs) or seemed responsible for major improvements in performance when doing preliminary experiments. Furthermore, we optimised hyperparameters for task 1 first, then for task 2, and finally for task 3. Thereby, we used insights gained from previous optimisation studies for choosing the hyperparameters to tune for the current task. For example, we noticed that 15 epochs of training are best for task 1 and task 2. Therefore, we did not tune the number of epochs for task 3.

Also due to time and resource restrictions, we did not perform full grid search. Instead, we randomly sampled 20 hyperparameter configurations for task 1 and task 3, and 30 hyperparameter configurations for task 2, using functionality provided by the Ray Tune library.

In each case, we train the model on the train data and evaluate it on the development data (see Sec.~\ref{sec:data}). In Sec.~\ref{sec:results:hyperparameters}, we report our main findings.

\begin{table}
\begin{subtable}{\linewidth}
\centering
\resizebox{\linewidth}{!}{
\begin{tabular}{lcc}
\toprule
     Hyperparameter & Values & Best value  \\
\midrule
     Batch size    & $\{16, 64\}$ & 16\\
     Epochs        & $\{15, 20, 25\}$ & 15\\ 
     Dropout       & $\{0.0, 0.1\}$ & 0.1\\
     Use LSTM      & $\{\textsc{True}, \textsc{False}\}$ & \textsc{True}\\
     Hidden dim    & $\{128, 256, 512\}$ & 512\\
     Embedding dim & $\{32, 64, 128\}$ & 128\\
     Transliteration & $\{\textsc{True}, \textsc{False}\}$ & \textsc{True}\\
\bottomrule
\end{tabular}
}
\caption{Hyperparameters for T1}
\label{tab:hyperparams:T1}
\end{subtable}

\begin{subtable}{\linewidth}
\centering
\resizebox{\linewidth}{!}{
\begin{tabular}{lcc}
\toprule
     Hyperparameter & Values & Best value  \\
\midrule
     Epochs        & $\{15, 20\}$ & 15 \\
     Hidden dim    & $\{256, 512\}$ & 512 \\
     Embedding dim & $\{32, 64, 128\}$ & 128 \\
     Transliteration & $\{\textsc{True}, \textsc{False}\}$ & \textsc{False}\\
     Rule freq. cutoff & $\{1, 5, 50\}$ & 1 \\
     Char2Token & $\{\textsc{lstm}, \textsc{max}\}$ & \textsc{max}\\
     Tag rules  & $\{\textsc{True}, \textsc{False}\}$ & \textsc{False} \\
\bottomrule
\end{tabular}
}
\caption{Hyperparameters for T2}
\label{tab:hyperparams:T2}
\end{subtable}

\begin{subtable}{\linewidth}
\centering
\resizebox{\linewidth}{!}{
\begin{tabular}{lcc}
\toprule
     Hyperparameter & Values & Best value  \\
\midrule
     Batch size    & $\{16, 64\}$ & 16 \\
     Hidden dim    & $\{256, 512\}$ & 512 \\
     Dropout       & $\{0.0, 0.1\}$ & 0.1 \\
     Transliteration & $\{\textsc{True}, \textsc{False}\}$ & \textsc{True}\\
     Char2Token & $\{\textsc{lstm}, \textsc{max}\}$ & \textsc{max}\\
\bottomrule
\end{tabular}
}
\caption{Hyperparameters for T3}
\label{tab:hyperparams:T3}
\end{subtable}
\caption{Tuned hyperparameters and corresponding values for all tasks.}
\label{tab:hyperparams}
\end{table}

\end{document}